\documentclass[runningheads]{llncs}

\usepackage{eccv}

\usepackage{eccvabbrv}

\usepackage{graphicx}
\usepackage{booktabs}
\usepackage{microtype}
\usepackage{xcolor}

\usepackage[accsupp]{axessibility}  

\usepackage{hyperref}

\usepackage{orcidlink}

\begin{document}

\title{\texorpdfstring{ATLASFusion: Aggregation Tracking with Location-Aware Sparse Fusion for Robust Spatio-Temporal Multi-View Pedestrian Tracking}{ATLASFusion: Aggregation Tracking with Location-Aware Sparse Fusion for Robust Spatio-Temporal Multi-View Pedestrian Tracking}}

\titlerunning{ATLASFusion}

\author{Keisuke Toida\inst{1}\orcidlink{0009-0006-4873-3651} \and
  Taigo Sakai\inst{1}\orcidlink{0009-0000-5130-1522} \and
  Takeshi Nakamura\inst{2}\orcidlink{0009-0001-4991-3383}\and
  Hiroshi Shimizu\inst{2}\orcidlink{0009-0001-1740-9123}\and
  Kazuhiro Hotta\inst{1}\orcidlink{0000-0002-5675-8713}}

\authorrunning{K.Toida et al.}

\institute{Meijo University, 1-501 Shiogamaguchi, Tempaku-ku, Nagoya 468-8502, Japan \email{190442098@ccalumni.meijo-u.ac.jp} \email{200442066@ccalumni.meijo-u.ac.jp} \email{kazuhotta@meijo-u.ac.jp} \and
  Chubu Electric Power Co., Inc., 1-1 Higashishin-cho, Higashi-ku, Nagoya 461-8680, Japan
  \email{Nakamura.Takeshi@chuden.co.jp} \email{Shimizu.Hiroshi@chuden.co.jp}\\}

\maketitle

\begin{abstract}
For multimedia spatial intelligence through time, multi-view multi-object tracking (MVMOT) suffers from persistent challenges in maintaining consistent object identities across different camera views, leading to tracking inaccuracies. A key source of these errors in modern methods is the distortion of feature representations when projecting from multiple views into a unified Bird's-Eye-View (BEV) space. This projection often creates non-uniform feature densities, harming the reliability of the fused representation. 
To address this, we present ATLASFusion, a reliability-aware sparse BEV fusion framework built on three complementary techniques: a sparse perspective transform that projects only valid feature points to prevent interpolation artifacts, density-aware weighted aggregation that assigns higher confidence to spatially reliable regions, and per-view BEV supervision that supervises each camera's BEV features independently before fusion. Experiments on the WildTrack and MultiViewX benchmarks demonstrated improvements in IDF1, MODP, and robustness over the baseline. ATLASFusion achieved a 95.9\% IDF1 tracking score on WildTrack, the highest among the compared methods and improved the localization precision (MODP) from 75.0\% to 89.2\% on MultiViewX. ATLASFusion also exhibited strong robustness under calibration noise. When high-magnitude noise was injected, TrackTacular failed entirely at 0.0\% MODA, whereas ATLASFusion retained 40.1\% MODA. When the input resolution was reduced by half, MVTr failed entirely and TrackTacular degraded by 2.7\% MODA, while ATLASFusion sustained 92.5\% MODA with only a 1.1\% decrease.
These gains were achieved with negligible computational overhead.

\end{abstract}

\section{Introduction}
\label{sec:intro}

Multi-object tracking (MOT) identifies and tracks multiple objects, such as pedestrians and vehicles, across frames in video sequences~\cite{simple, simplemetric}. Its broad applicability in surveillance, robotic vision, autonomous driving, and sports analytics has driven sustained research interest. This makes MVMOT a central task for multimedia spatial intelligence through time.

In multi-view multi-object tracking (MVMOT), the same object observed from different viewpoints is associated and its movement is tracked across a larger area. However, challenges unique to the multi-view setting persist, including appearance variations due to changes in camera angle and lighting conditions, occlusions caused by blind spots, and increased computational overhead from simultaneous multi-camera processing.

\begin{figure}[t]
  \centering
  \includegraphics[width=0.95\linewidth]{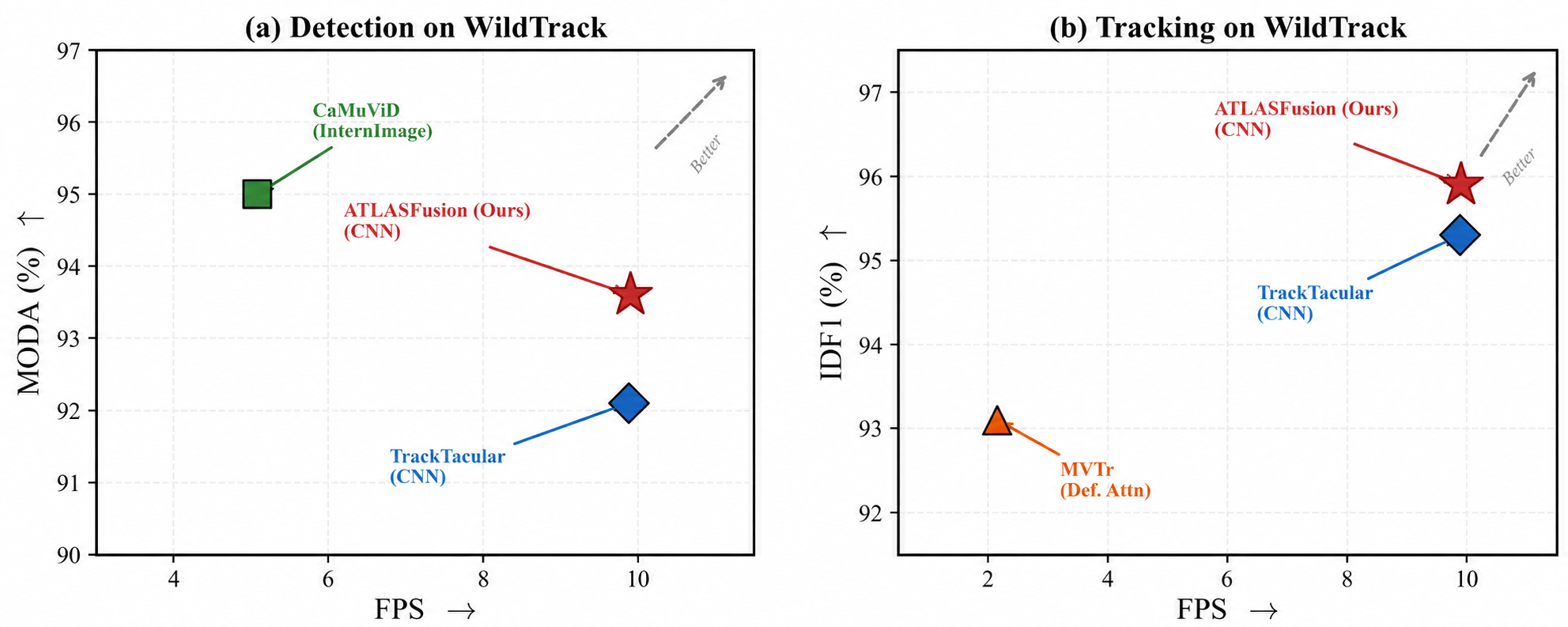}
  \caption{Comparison of accuracy and computational efficiency on the WildTrack dataset. (a) Detection accuracy (MODA) and (b) tracking accuracy (IDF1) are plotted against inference speed (FPS). ATLASFusion archives the highest IDF1 score among the methods shown while maintaining competitive detection accuracy and throughput.}
  \label{fig:intro}
\end{figure}

MVMOT leverages multiple viewpoints to enhance robustness against occlusion. However, projecting features into a common Bird's-Eye-View (BEV)~\cite{bevproj} space introduces additional challenges. The non-uniform density of projected features, caused by variations in object scale with distance, remains largely unaddressed. Such density distortion degrades detection accuracy, particularly for distant objects. This paper addresses the problem of feature non-uniformity in BEV projection.

Three solutions are proposed: (1) a sparse transformation method that avoids unnatural pixel interpolation, (2) density-aware weighting that adjusts feature fusion based on spatial confidence and distance to cameras, and (3) per-view BEV supervision that encourages each camera's features to be individually informative for BEV detection.
Unlike previous multi-view approaches that rely solely on a unified BEV-level loss,
the proposed method explicitly introduces a \textit{self-view discriminability supervision} that enables each camera view to learn detection-friendly BEV representations prior to fusion.


The robust performance of ATLASFusion was validated on standard benchmarks. A 95.9\% IDF1 tracking score, the highest among the compared methods, was achieved on the WildTrack dataset, and MOTP on the MultiViewX dataset was improved to 85.4\%.
\cref{fig:intro} illustrates the accuracy--efficiency trade-off of recent methods on the WildTrack dataset. ATLASFusion attained the highest IDF1 score among the methods shown while maintaining real-time inference speed and competitive MODA, demonstrating that the proposed modules introduce negligible computational overhead.

The contributions of this paper are as follows.
\begin{itemize}
  \item We formulate robust multi-view pedestrian tracking as reliability-aware BEV spatial intelligence for constructing stable spatio-temporal representations from multi-camera video.
  \item The proposed Sparse Perspective Transform projects only valid feature points into BEV space, bypassing the dense warping that stretches features. This Sparse Perspective Transform reduces interpolation-induced feature distortion.
  \item Density-Aware Fusion assigns higher weights to spatially reliable features, such as those closer to the camera. The resulting fused BEV map reduces the missing or attenuated regions that arise from naive feature averaging.
  \item Per-View BEV Supervision supervises each camera view to produce correct BEV detections independently before fusion. This per-view discriminability improves the final fusion quality without directly coupling predictions across views.
\end{itemize}


\section{Related works}
\label{sec:related}



\subsection{Multi-View Multi-Object Tracking (MVMOT)}
\label{sec:related:mvmot}

%


%

MVMOT tracks multiple objects from multiple viewpoints. 
Compared to single-view methods, MVMOT mitigates issues such as occlusion and blind spots by leveraging information from multiple perspectives. MVMOT approaches are categorized into two types based on the stage at which multi-view information is fused: Late Multi-View and Early Multi-View.
The Late Multi-View approach performs detection independently on each camera and then fuses the results. Representative examples include PolarMOT~\cite{polarmct}, which projects 2D detection results into a BEV polar coordinate system, and MTMCT~\cite{mtmct}, which performs re-identification based on features from single-view tracking. However, this paradigm is limited because information from each camera remains separate during detection; a detection error in one view due to occlusion cannot be corrected by another view, thereby constraining overall accuracy.

In contrast, the Early Multi-View approach aggregates feature maps from all cameras into a unified BEV before performing detection. By sharing information across all cameras at the feature level, this approach is more robust to occlusion and achieves more accurate object localization. Methods such as MVDet~\cite{mvdet}, MUTR3D~\cite{mutr}, and BEVFormer~\cite{bevformer} have demonstrated that the Early approach yields superior localization accuracy. EarlyBird~\cite{early} extended the MVDet architecture by projecting camera features to BEV and performing joint detection and tracking in an end-to-end manner through a unified network. TrackTacular~\cite{tacular} also follows the Early Multi-View paradigm, fusing camera features in BEV space to achieve high performance in both detection and tracking.
After aggregating BEV-projected features from all cameras, TrackTacular stores the BEV feature map from the previous frame, denoted as History BEV($t{-}1$), before the detection decoder and reuses it at frame $t$, effectively addressing occlusion.

An alternative line of research exploits longer temporal histories to improve tracking accuracy. MVTrajecter~\cite{mvtrajecter} aggregated features and positional information from multiple past frames (e.g., the most recent seven frames) and computed inter-frame relationships through attention mechanisms, achieving high MOTA and IDF1 scores. However, retaining multi-frame features incurs substantial GPU memory consumption and reduces inference speed (FPS). In contrast, one-shot tracking methods such as TrackTacular~\cite{tacular} utilize only the current frame and a single adjacent past frame, thereby maintaining low computational overhead and real-time capability.
A clear trade-off therefore exists between the accuracy gains afforded by long-term trajectory information and the processing speed as well as resource requirements imposed by retaining that history.
Furthermore, the unified cost computation in MVTrajecter relies on a hyperparameter $\alpha$ that balances the trajectory motion cost and the trajectory appearance cost. Optimal tracking performance requires this weighting parameter to be tuned empirically according to the camera installation environment, crowd density, and dataset characteristics, which limits the generalizability of the method across diverse real-world settings.

Despite these advances, a fundamental challenge remains: the projection from 2D to BEV creates feature distortion and non-uniform density. This problem is particularly pronounced as the scale of objects changes with distance, degrading the final detection quality.
The proposed ATLASFusion directly addresses this projection problem to further enhance MVMOT performance.
Although existing early fusion methods such as MVDet, EarlyBird, and TrackTacular jointly optimize the fused BEV features across multiple views, none of them supervise each single-view BEV independently.
As a result, individual camera features are less discriminative before fusion, which limits their contribution to the final BEV representation.
ATLASFusion explicitly addresses this limitation through single-view BEV supervision.

\section{Proposed Method}

\subsection{Overview of the Proposed Method: ATLASFusion}

\begin{figure}[t]
  \centering
  \includegraphics[width=0.90\linewidth]{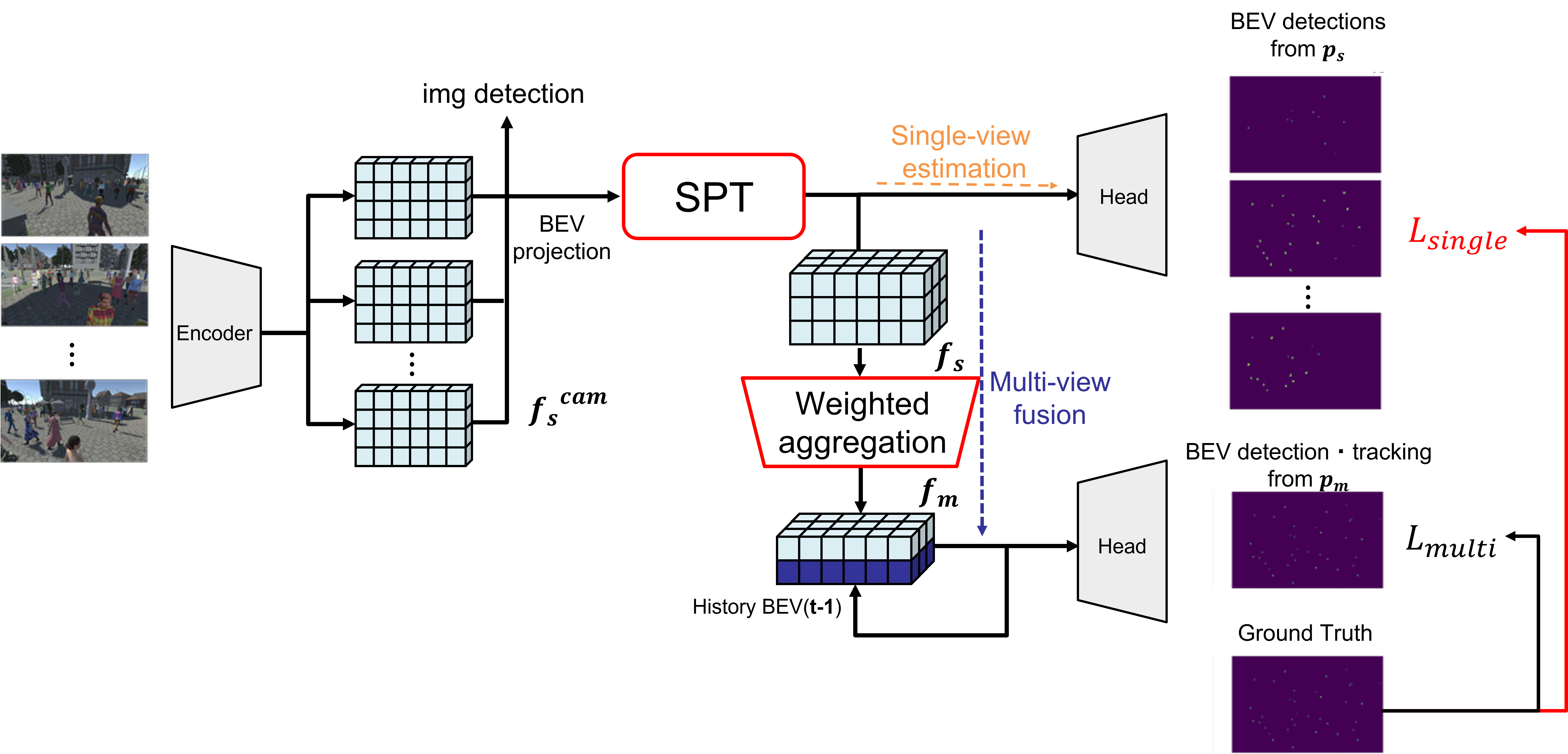}
  \caption{Overview of the ATLASFusion. Each camera view is projected sparsely into BEV space, followed by confidence-based fusion and per-view supervision.}
  \label{fig:overview}
\end{figure}

The overall architecture of the proposed ATLASFusion is illustrated in \cref{fig:overview}.
Features are first extracted from each camera image and then split into two branches: a single-view estimation branch and a multi-view fusion branch.
The upper branch performs per-camera density estimation, providing direct supervision for each view before multi-view fusion.
In the lower branch, features from all cameras are projected into a shared Bird's-Eye View (BEV) space, where viewpoint differences and missing observations are compensated to produce the final fused density map.
The training objective combines the losses from both branches,
enabling the model to preserve single-view accuracy while retaining detection-relevant information in each view.
Features extracted from multiple cameras are transformed through three key components.

\begin{figure}[t]
  \centering
  \includegraphics[width=0.65\linewidth]{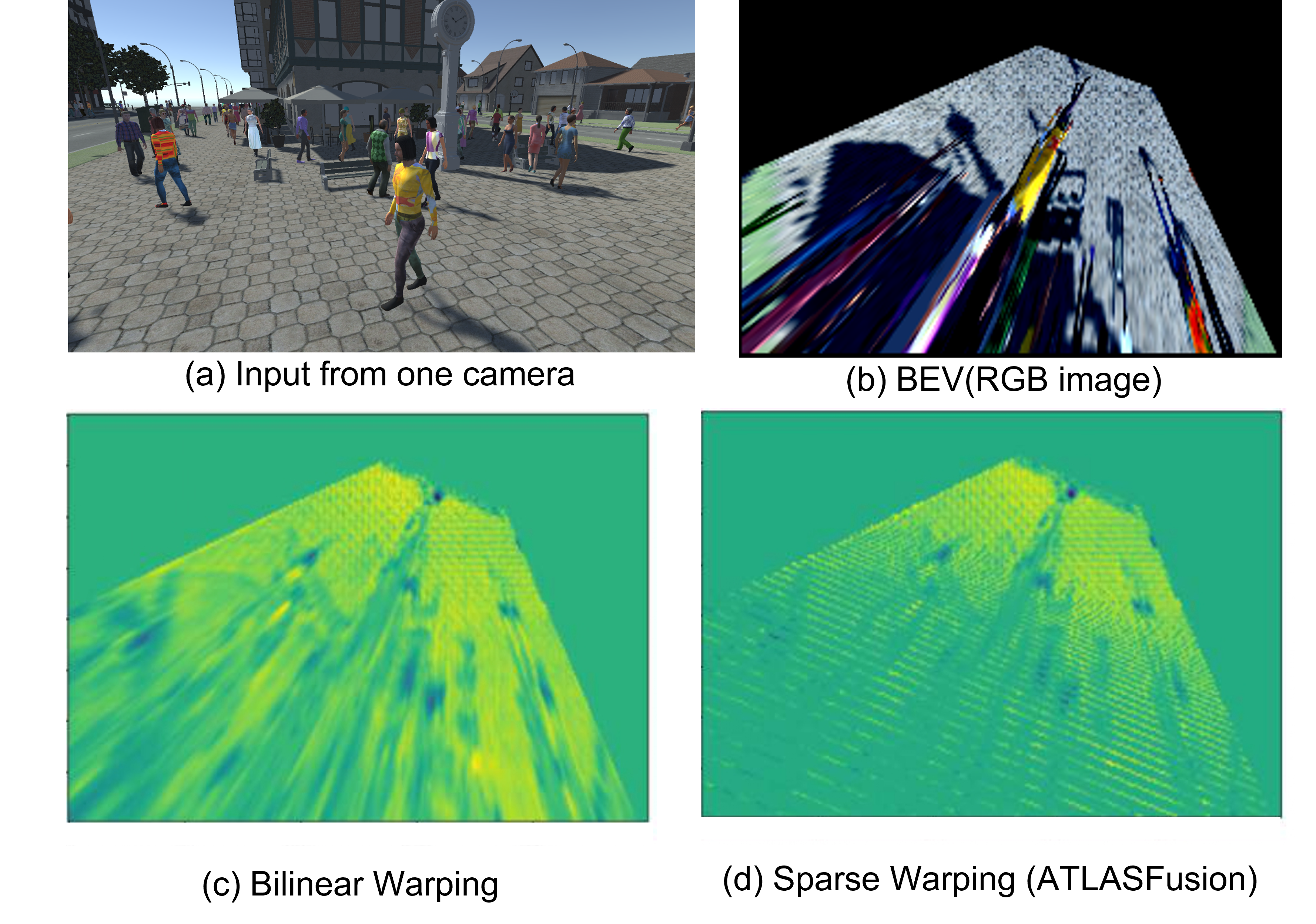}
  \caption{
    (a) Original perspective RGB image. (b) Simple BEV projection, showing strong stretching and distortion.
    Bottom of row shows the comparison of two warping methods. (c) Bilinear warping. (d) Sparse warping by our Sparse Perspective Transform (SPT).}
    \label{fig:warp_comparison}
\end{figure}

\begin{figure}[t]
  \centering
  \includegraphics[width=0.85\linewidth]{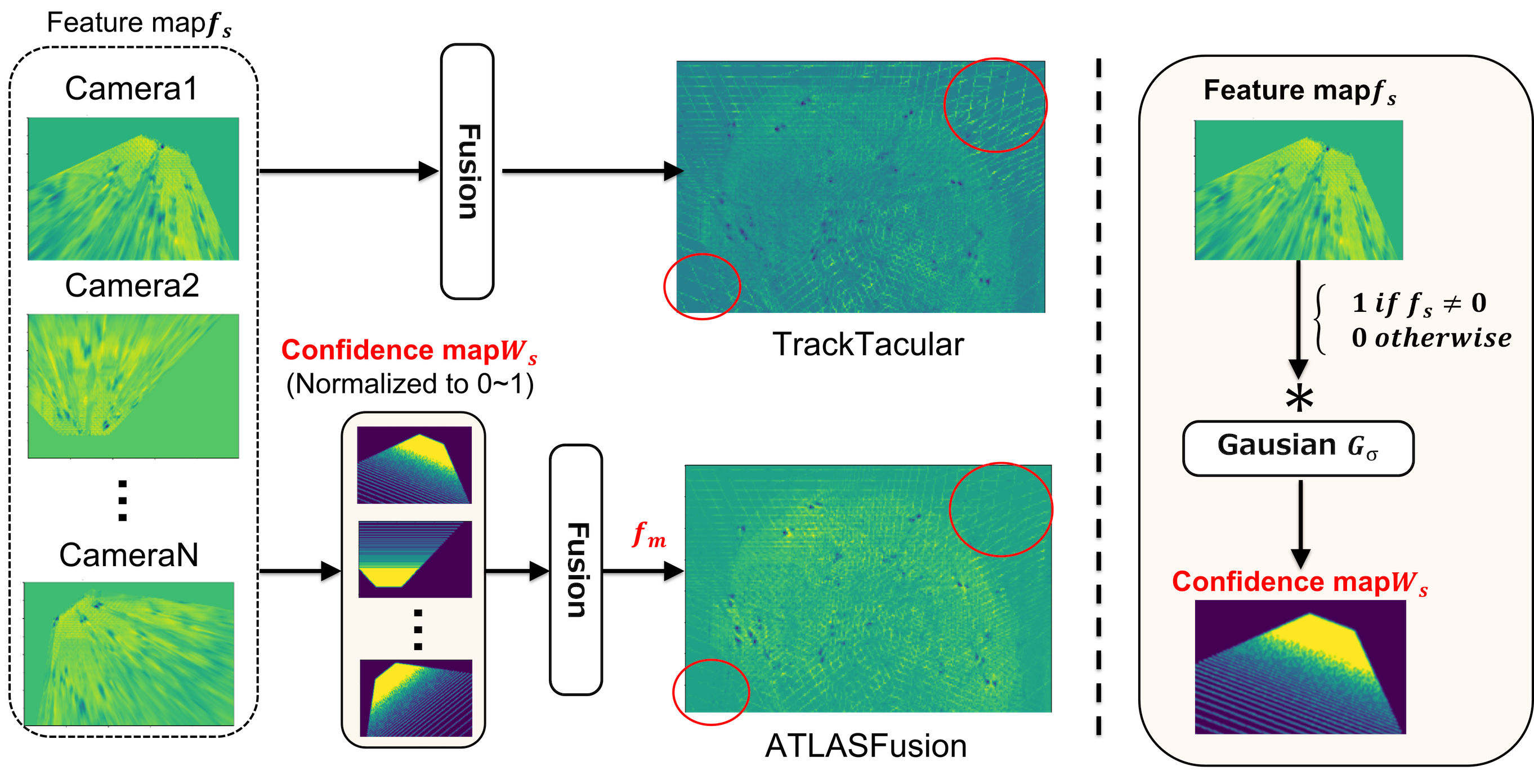}
  \caption{Overview of Weighted Aggregation}
  \label{fig:ours_1_2}
\end{figure}

\begin{enumerate}
  \item \textbf{Sparse Perspective Transform (SPT):}
        Dense warping stretches nearby features and blurs distant ones in BEV space. As shown in \cref{fig:warp_comparison}, Dense warping in (c) stretches each pedestrian into a long radial streak, making individual shapes indistinguishable. Our proposed sparse projection in (d) keeps each pedestrian as a compact activation cluster, so their BEV footprint remains separable instead of collapsing into a blurred surface.

  \item \textbf{Density-Aware Weighted Aggregation:}
        In each camera view, features from distant areas become sparse after BEV projection,
        making those regions less reliable when simply averaged.
        As shown in \cref{fig:ours_1_2}, the Simple fusion used in TrackTacular as shown in top right produces clear missing regions in red circles, where distant features vanish after averaging. In contrast, our density-aware weighting as shown in bottom right fills these gaps and yields a continuous feature field across the BEV space.

  \item \textbf{Per-View BEV Supervision:} We compute a detection loss ($L_{\mbox{single}}$) on the single-view BEV features before fusion and combine it with the loss ($L_{\mbox{multi}}$) from the final fused features. This encourages each view's features to be independently discriminative, thereby improving overall robustness.
\end{enumerate}





\subsection{Sparse Perspective Transform}

The problem of \textit{information density distortion} in the projection from camera images to BEV space has not been adequately addressed by bilinear interpolation-based methods. In perspective images, nearby objects are represented with higher pixel density than distant ones.
When projected into BEV space, sparse regions representing nearby objects are unnaturally stretched.
Such stretching causes spatial misalignment and distorts the feature representation, degrading the ability to accurately localize objects.

To address this, a \textbf{sparse projection method} was introduced that selectively projects only valid feature points, preserving the natural density distribution of the scene.
For each BEV grid location $\mathbf{q}$, the planar homography maps it to the corresponding location $\mathbf{u}_s(\mathbf{q})$ in the camera feature map.
SPT obtains the feature from the nearest discrete location without bilinear interpolation.

\begin{equation}
  f_s(\mathbf{q}) =
  \left\{
  \begin{array}{ll}
    f_s^{cam}\!\left(\mathrm{round}(\mathbf{u}_s(\mathbf{q}))\right)
    & \mbox{if } \mathbf{u}_s(\mathbf{q}) \in \Omega_s,\\
    0 & \mbox{otherwise},
  \end{array}
  \right.
  \label{eq:spt}
\end{equation}
where $\Omega_s$ denotes the valid region of the camera feature map, and
$\mathrm{round}(\cdot)$ rounds the continuous projected coordinate to the nearest integer grid location.
Unlike dense bilinear warping, SPT does not interpolate features from neighboring locations.

\cref{fig:warp_comparison} presents a comparison of the input view and its projected BEV representations. \cref{fig:warp_comparison}(c) and (d) compare the proposed sparse warping method with conventional bilinear warping.

As shown in \cref{fig:warp_comparison}(c), bilinear warping stretches distant features and blurs object boundaries,
because it interpolates features continuously along viewing rays without considering depth changes.
In contrast, our sparse projection in \cref{fig:warp_comparison}(d) keeps feature points sharp and geometrically aligned,
preserving the natural perspective density of the scene.


\subsection{Density-Aware Weighted Aggregation}

Higher confidence is assigned to dense (nearby) regions and lower confidence to sparse (distant) ones.
"Density" here reflects how feature projections spread over the ground depending on camera distance.
In perspective projection, nearby regions appear with higher sampling density, while distant areas are projected more sparsely because each pixel covers a larger ground area.


The effectiveness of our weighted aggregation is illustrated in \cref{fig:ours_1_2}.
In the simple fusion baseline (TrackTacular), regions with missing observations are averaged together, which produces falsely amplified responses in areas where no valid features should exist as highlighted in red in \cref{fig:ours_1_2}.
In contrast, ATLASFusion assigns weights based on confidence,
suppressing these artificial peaks and generating a smoother and physically reliable fused feature map.

Let $i_s \in \mathbf{R}^{C \times H \times W}$ denote the input image from the $s$-th camera, and let $E(\cdot)$ denote the feature extractor. The camera-space feature map $f^{cam}_s$ is obtained as
$f^{cam}_s = E(i_s).$
Then, we transform $f^{cam}_s$ into BEV feature $f_s$ through a geometric projection $T(\cdot)$ that maps each feature to its corresponding ground-plane coordinate as $f_s = T(f^{cam}_s, K_s, R_s, \mathbf{t_s})$, 
where $T(\cdot)$ represents the perspective-to-BEV mapping derived from the calibrated camera parameters $(K_s, R_s, \mathbf{t_s})$.
$K_s$, $R_s = [\mathbf{r}_1, \mathbf{r}_2, \mathbf{r}_3]$, and $\mathbf{t_s}$ denote the intrinsic matrix, the rotation matrix, and the translation vector of the $s$-th camera, respectively.
Since all target objects are assumed to lie on the ground plane ($Z=0$), full 3D-to-2D projection reduces to a planar homography. For a ground-plane point $\mathbf{X} = [x, y, 0, 1]^\top$, the third column of $R_s$ corresponding to the $Z$-axis is eliminated, yielding the $3\times 3$ homography matrix as
$H_s = K_s [\mathbf{r}_1, \mathbf{r}_2, \mathbf{t_s}]$.
Given a pixel coordinate $\mathbf{u}$, the corresponding BEV coordinate $\mathbf{q} = [x, y, 1]^\top$ is recovered by the inverse homography up to a scale factor $\lambda$, $\lambda\,\mathbf{q} = H_s^{-1}\,\mathbf{u}$.
Note that $\mathbf{q}$ denotes BEV coordinate,
then $f_s(\mathbf{q})$ represents the feature value assigned to that position after projection.

After obtaining the BEV feature $f_s$ for each camera,
the proposed sparse projection strategy is applied to perform the perspective-to-BEV mapping more faithfully.
Instead of dense bilinear sampling, only valid features are mapped. A feature point is considered invalid if it falls outside the predefined BEV grid.
The validity mask is determined from the projected coordinate rather than from the numerical feature value.
Thus, a sparse mask $M_s$ is generated and applied to a \textit{Gaussian filter} $G_\sigma$ to produce a confidence map $W_s$ as 
\begin{equation}
  M_s(\mathbf{q}) = \left\{
  \begin{array}{ll}
    1 & \mbox{if } \mathbf{u}_s(\mathbf{q}) \in \Omega_s \\
    0 & \mbox{otherwise}
  \end{array}
  \right.
  , \quad W_s = G_\sigma * M_s
\end{equation}

This filtering step allows the confidence to spread smoothly around valid feature points, assigning graduated weights to neighboring pixels rather than creating a hard boundary.
The resulting smooth representation enables more robust learning from sparse features.
Applying the Gaussian filter to the binary mask $M_s$ fills small gaps between valid points, yielding a smooth confidence map $W_s$.
This map $W_s$ encodes the spatial reliability of each BEV position and serves as the weighting map when fusing features from multiple cameras. 
SPT prevents feature spreading caused by bilinear interpolation, while $W_s$ compensates for differences in feature density across BEV regions.

The \textbf{multi-view fused feature} $f_m$ is obtained by confidence-weighted summation over all camera views, aggregating per-view features while emphasizing those with higher confidence as
\begin{equation}
  f_m = \sum_{s=1}^S (f_s \cdot W_s)
  \label{eq:weighted_fusion}
\end{equation}
where $S$ denotes the number of cameras.
This assigns high confidence to dense and reliable regions while reducing the influence of sparse or uncertain areas.

\subsection{Per-View BEV Supervision}

To ensure that each camera's BEV feature learns independently useful representations for detection, \textbf{per-view BEV supervision} was introduced. This auxiliary loss supervises each camera independently before fusion, rather than directly enforcing consistency between camera pairs.

The fused feature $f_m$ is formed by the confidence-weighted aggregation in \cref{eq:weighted_fusion}. Detection is performed in BEV space using features at current and previous frames for temporal consistency. Let $D(\cdot)$ denote the detection head; detection probabilities are computed for both the single-view BEV features and final fused multi-view features. Let $p_s$ denote the detection probability map from the $s$-th single-view feature $f_s$,
and $p_m$ denote the probability map from the fused multi-view feature $f_m$. These maps are computed using features at the current and previous frames for temporal consistency
\begin{eqnarray}
  p_s &= D(f_s, f^{(t-1)}_s), \quad p_m &= D(f_m, f^{(t-1)}_m).
\end{eqnarray}

The \textit{Focal Loss}~\cite{DBLP:journals/corr/abs-1708-02002}, which is robust to class imbalance, was adopted. The loss is computed only at valid projected locations, defined as the regions on the BEV that are visible from at least one camera view
\begin{eqnarray}
  L_{\mbox{single}} &=& -\sum_{s=1}^S \sum_i \alpha_i (1 - p_{i,s})^\gamma \big[ y_{i,s} \log p_{i,s}
    + (1 - y_{i,s}) \log (1 - p_{i,s}) \big]
\end{eqnarray}
\begin{eqnarray}
  L_{\mbox{multi}} &=& -\sum_i \alpha_i (1 - p_i)^\gamma [y_i \log p_i + (1 - y_i) \log (1 - p_i)]
\end{eqnarray}
where $\alpha_i$ is the class-balancing weight, $\gamma$ is the focusing parameter. As illustrated in \cref{fig:overview}, our training objective combines the detection loss from single-view features ($L_{\mbox{single}}$) with the loss from the fused multi-view features ($L_{\mbox{multi}}$).
The final detection loss $L_{\mbox{det}}$ is defined as

\begin{equation}
  L_{\mbox{det}} = \beta L_{\mbox{single}} + L_{\mbox{multi}}
\end{equation}
where $\beta$ is a balance coefficient.
Unlike prior works that mainly optimize the fused BEV~\cite{mvdet, mvdetr}, each camera's single-view BEV is directly supervised with a detection loss before fusion. Detection probabilities from both single-view BEV features and the fused BEV are computed and optimized jointly to provide auxiliary per-view supervision
and improve robustness.
This loss function enables each camera's BEV features to learn detection-friendly representations independently, prior to multi-view fusion.
$\alpha, \gamma, \beta$ are specified in Section~4.1.

\section{Experiments}

\begin{figure}[ht]
  \centering
  \includegraphics[width=0.95\linewidth]{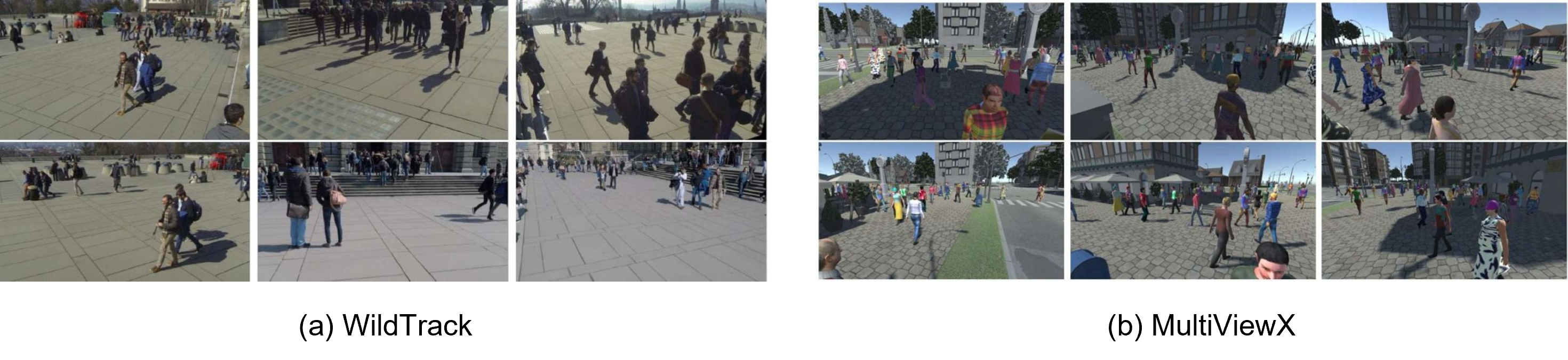}
  \caption{(a) WildTrack dataset and (b) MultiViewX dataset.}
  \label{fig:dataset}
\end{figure}

\subsection{Experimental Setup}

ATLASFusion was evaluated on the WildTrack dataset~\cite{ksp} and the MultiViewX dataset~\cite{hou2020multiview}.
WildTrack is a large-scale multi-camera pedestrian detection and tracking dataset captured in a real-world outdoor environment. As shown in \cref{fig:dataset}(a), it comprises 7 static HD cameras with overlapping fields of view, each with a resolution of $1920 \times 1080$ pixels and a synchronized frame rate of 2 FPS. The dataset contains 400 annotated frames, totaling 2800 images across all views. The average number of pedestrians per frame is approximately 24.

MultiViewX is a synthetic multi-view pedestrian detection and tracking dataset rendered by a game engine. As shown in \cref{fig:dataset}(b), it comprises six virtual cameras with overlapping fields of view.
The average number of pedestrians per frame is approximately 40. Ground-truth annotations are provided for both detection and tracking in the BEV space ($640 \times 1000$ pixels), enabling precise evaluation of detection and tracking in world coordinates.

For the Gaussian filter used in density estimation, the standard deviation was set to $\sigma = 1$. The parameters for the Focal Loss were set to $\alpha = 2$ and $\gamma = 4$, following the same configuration as the baseline TrackTacular. The balancing coefficient $\beta$ for per-view BEV supervision was set to 0.1, which yielded the best validation accuracy among tested values ${0.001,\ 0.1,\ 1.0,\ 10,\ 100}$.
All experiments were conducted on a PC equipped with an RTX 4080 Super GPU and an Intel i7-14700K CPU.

Quantitative evaluation was conducted on the BEV detection map using standard 2D detection and tracking metrics. MODA measures detection accuracy by considering false positives and false negatives. MODP evaluates the average localization precision of correct detections. MOTA reflects overall tracking accuracy, accounting for false positives, false negatives, and ID switches. MOTP indicates the average localization error of matched objects. IDF1 is the F1 score based on correctly identified detections.



\subsection{Results and Analysis}

\begin{table}[t]
  \centering
  \setlength{\tabcolsep}{3.0pt}
  \scalebox{0.9}{
    \begin{tabular}{rcccc}
      \toprule
                                     & MODA             & MODP             & Precision        & Recall           \\
      \midrule
      RCNN \& Cluster~\cite{cluster} & 11.3             & 18.4             & 68               & 43               \\
      DeepMCD~\cite{deeepmcd}        & 67.8             & 64.2             & 85               & 82               \\
      Deep-Occlusion~\cite{deepocc}  & 74.1             & 53.8             & 95               & 80               \\
      \midrule
      MVDet~\cite{mvdet}             & 88.2             & 75.7             & 94.7             & 93.6             \\
      SHOT~\cite{shot}               & 90.2             & 76.5             & 96.1             & 94.0             \\
      MVDeTr~\cite{mvdetr}           & 91.5             & \underline{82.1} & \underline{97.4} & 94.0             \\
      Booster-SHOT~\cite{booster}    & 92.8             & \textbf{84.9}    & \textbf{97.5}    & 95.3             \\
      EarlyBird~\cite{early}         & 91.2             & 81.8             & 94.9             & \underline{96.3} \\
      TrackTacular~\cite{tacular}    & 92.1             & 76.2             & 97.0             & 95.1             \\
      {CaMuViD~\cite{camuvid}}       & {\textbf{95.0}}  & {80.5}           & {95.8}           & {\textbf{99.2}}  \\
      \midrule
      ATLASFusion (Ours)             & \underline{93.6} & \underline{82.1} & \underline{97.4} & 96.2             \\
      \bottomrule
    \end{tabular}}
  \caption{Detection performance on the WildTrack dataset. All metrics are in \%, where $\uparrow$ indicates higher is better. The bold numbers indicate the best accuracy in each column, while the underlined ones represent the second best.}
  \label{tab:detection-sota-wild}
\end{table}

\begin{table}[t]
  \centering
  \setlength{\tabcolsep}{3.0pt}
  \scalebox{0.9}{
    \begin{tabular}{rcccc}
      \toprule
                                     & MODA             & MODP             & Precision        & Recall           \\
      \midrule
      RCNN \& Cluster~\cite{cluster} & 18.7             & 46.4             & 63.5             & 43.9             \\
      DeepMCD~\cite{deeepmcd}        & 70.0             & 73.0             & 85.7             & 83.3             \\
      Deep-Occlusion~\cite{deepocc}  & 75.2             & 54.7             & 97.8             & 80.2             \\
      \midrule
      MVDet~\cite{mvdet}             & 83.9             & 79.6             & 96.8             & 86.7             \\
      SHOT~\cite{shot}               & 88.3             & 82.0             & 96.6             & 91.5             \\
      MVDeTr~\cite{mvdetr}           & 93.7             & \underline{91.3} & \underline{99.5} & 94.2             \\
      Booster-SHOT~\cite{booster}    & \underline{94.4} & \textbf{92.0}    & 99.4             & 94.9             \\
      EarlyBird~\cite{early}         & 94.2             & 90.1             & 98.6             & 95.7             \\
      TrackTacular~\cite{tacular}    & \textbf{96.5}    & 75.0             & 99.4             & \textbf{97.1}    \\
      \midrule
      ATLASFusion (Ours)             & \textbf{96.5}    & 89.2             & \textbf{99.8}    & \underline{96.7} \\
      \bottomrule
    \end{tabular}}
  \caption{Detection performance on the MultiViewX dataset. All metrics are in \%, where $\uparrow$ indicates higher is better. The bold numbers indicate the best accuracy in each column, while the underlined ones represent the second best.}
  \label{tab:detection-sota-multi}
\end{table}

The performance of ATLASFusion was benchmarked against state-of-the-art methods on the WildTrack and MultiViewX datasets, with results presented in \cref{tab:detection-sota-wild}, \cref{tab:detection-sota-multi}, and \cref{tab:track}.
\cref{tab:detection-sota-wild} and \cref{tab:detection-sota-multi} summarize the detection performance of ATLASFusion and conventional multi-view methods on WildTrack and MultiViewX, respectively. Each table reports standard evaluation metrics for multi-view pedestrian detection: MODA and MODP evaluate overall and localization accuracy, while Precision and Recall measure detection reliability. The best and second-best results are highlighted in bold and underlined, respectively. ATLASFusion achieved a substantial improvement in MODP over the TrackTacular baseline, which reflects the precision of object localization.

\cref{tab:track} summarizes the tracking performance on the WildTrack and MultiViewX datasets.
The metrics include MOTA and MOTP for overall tracking accuracy and localization precision~\cite{MOT, eval}, and IDF1 for identity consistency. MT and ML denote the total count of mostly tracked and mostly lost objects during their lifespan, respectively.
All scores are expressed as percentages; higher values indicate better performance except for ML.
Bold and underlined numbers denote the best and second-best results, respectively.
ATLASFusion achieved competitive or superior tracking accuracy compared to previous methods, maintaining high identity consistency and stable trajectories.



\begin{table}[t]
  \setlength{\tabcolsep}{0.55mm}
  \centering
  \scalebox{0.9}{
    \begin{tabular}{rcccccc}
      \toprule
                                          & \multicolumn{5}{c}{Wildtrack}                                                                                 \\\cmidrule(lr){2-6}
                                          & IDF1$\uparrow$                 & MOTA$\uparrow$   & MOTP$\uparrow$     & MT$\uparrow$     & ML$\downarrow$    \\
      \midrule
      \small{KSP-DO~\cite{ksp}}           & 73.2                           & 69.6             & 61.5               & 28.7             & 25.1              \\
      \small{KSP-DO-ptrack~\cite{ksp}}    & 78.4                           & 72.2             & 60.3               & 42.1             & 14.6              \\
      \small{DMCT ~\cite{dmct}}           & 77.8                           & 72.8             & 79.1               & 61.0             & 4.9               \\
      \small{DMCT Stack ~\cite{dmct}}     & 81.9                           & 74.6             & 78.9               & 65.9             & \underline{4.9}   \\
      \small{GLMB-YOLOv3~\cite{glmb}}     & 74.3                           & 69.7             & 73.2               & 79.5             & 21.6              \\
      \small{ReST ~\cite{rest}}           & 86.7                           & 84.9             & 84.1               & \underline{87.8} & 4.9               \\\midrule
      \small{EarlyBird~\cite{early}}      & 92.3                           & 89.5             & 86.6               & 78.0             & \underline{4.9}   \\
      \small{TrackTacular~\cite{tacular}} & 95.3                           & 91.8             & 85.4               & \underline{87.8} & \underline{4.9}   \\
      {\small{MVTr~\cite{mvtr}}}          & {93.1}                         & {92.3}           & {\underline{92.7}} & {\textbf{95.1}}  & {\underline{4.9}} \\
      \small{MCBLT~\cite{mcblt}}          & \underline{95.6}               & \textbf{92.6}    & \textbf{93.7}      & 80.5             & 7.3               \\
      \midrule
      \small{ATLASFusion} (Ours)          & \textbf{95.9}                  & \underline{92.4} & 86.3               & 85.4             & \textbf{4.8}      \\

      \midrule
                                          & \multicolumn{5}{c}{MultiViewX}                                                                                \\\cmidrule(lr){2-6}
                                          & IDF1$\uparrow$                 & MOTA$\uparrow$   & MOTP$\uparrow$     & MT$\uparrow$     & ML$\downarrow$    \\
      \midrule
      \small{EarlyBird~\cite{early}}      & 82.4                           & 88.4             & \textbf{86.2}      & 82.9             & \textbf{1.3}      \\
      \small{TrackTacular~\cite{tacular}} & \textbf{85.6}                  & \underline{92.4} & 80.1               & \textbf{92.1}    & \underline{2.6}   \\
      \midrule
      \small{ATLASFusion} (Ours)          & \underline{85.0}               & \textbf{92.5}    & \underline{85.4}   & \underline{85.5} & 2.9               \\

      \bottomrule
    \end{tabular}}
  \caption{Tracking performance on the WildTrack and MultiViewX datasets. All metrics are in \%, where $\uparrow$ indicates higher is better. The bold numbers indicate the best accuracy in each column, while the underlined ones represent the second best.}

  \label{tab:track}
\end{table}

\begin{table}[t!]
  \setlength{\tabcolsep}{1.2pt}
  \centering
  \scalebox{0.9}{
    \begin{tabular}{lccccc|ccccc}
      \toprule
                          & \multicolumn{5}{c|}{WildTrack} & \multicolumn{5}{c}{MultiViewX}                                                                                                                                                                \\\cmidrule(l){2-6}\cmidrule(r){7-11}
                          & \multicolumn{2}{c}{Detection}  & \multicolumn{3}{c|}{Tracking}  & \multicolumn{2}{c}{Detection} & \multicolumn{3}{c}{Tracking}                                                                                                 \\\cmidrule(lr){2-3}\cmidrule(l){4-6}\cmidrule(r){7-8}\cmidrule(lr){9-11}
                          & MODA                           & MODP                           & IDF1                          & MOTA                         & MOTP          & MODA          & MODP          & IDF1          & MOTA          & MOTP          \\
      \midrule
      \small{Baseline}    & 92.1                           & 76.2                           & 95.3                          & 91.8                         & 85.4          & 96.5          & 75.0          & \textbf{85.6} & 92.4          & 80.1          \\
      \midrule
      $+$ \small{SPT}     & 93.0                           & 79.5                           & 95.5                          & 92.1                         & 86.4          & 96.3          & 84.7          & 85.7          & 92.4          & 83.2          \\
      $+$ \small{WA}      & 93.4                           & 80.9                           & 95.8                          & \textbf{92.5}                & \textbf{86.6} & 96.5          & 87.0          & 85.8          & \textbf{92.5} & \textbf{84.0} \\
      $+$ \small{PV loss} & \textbf{93.6}                  & \textbf{82.1}                  & \textbf{95.9}                 & 92.4                         & 86.3          & \textbf{96.5} & \textbf{89.2} & 85.0          & 92.5          & 85.4          \\
      \bottomrule
    \end{tabular}}
  \caption{Ablation study on WildTrack and MultiViewX. WA is Weight Aggregation.}
  \label{tab:model-ablation-wild}
  \label{tab:model-ablation-multi}
\end{table}

\begin{figure}[t]
  \centering
  \includegraphics[width=0.85\linewidth]{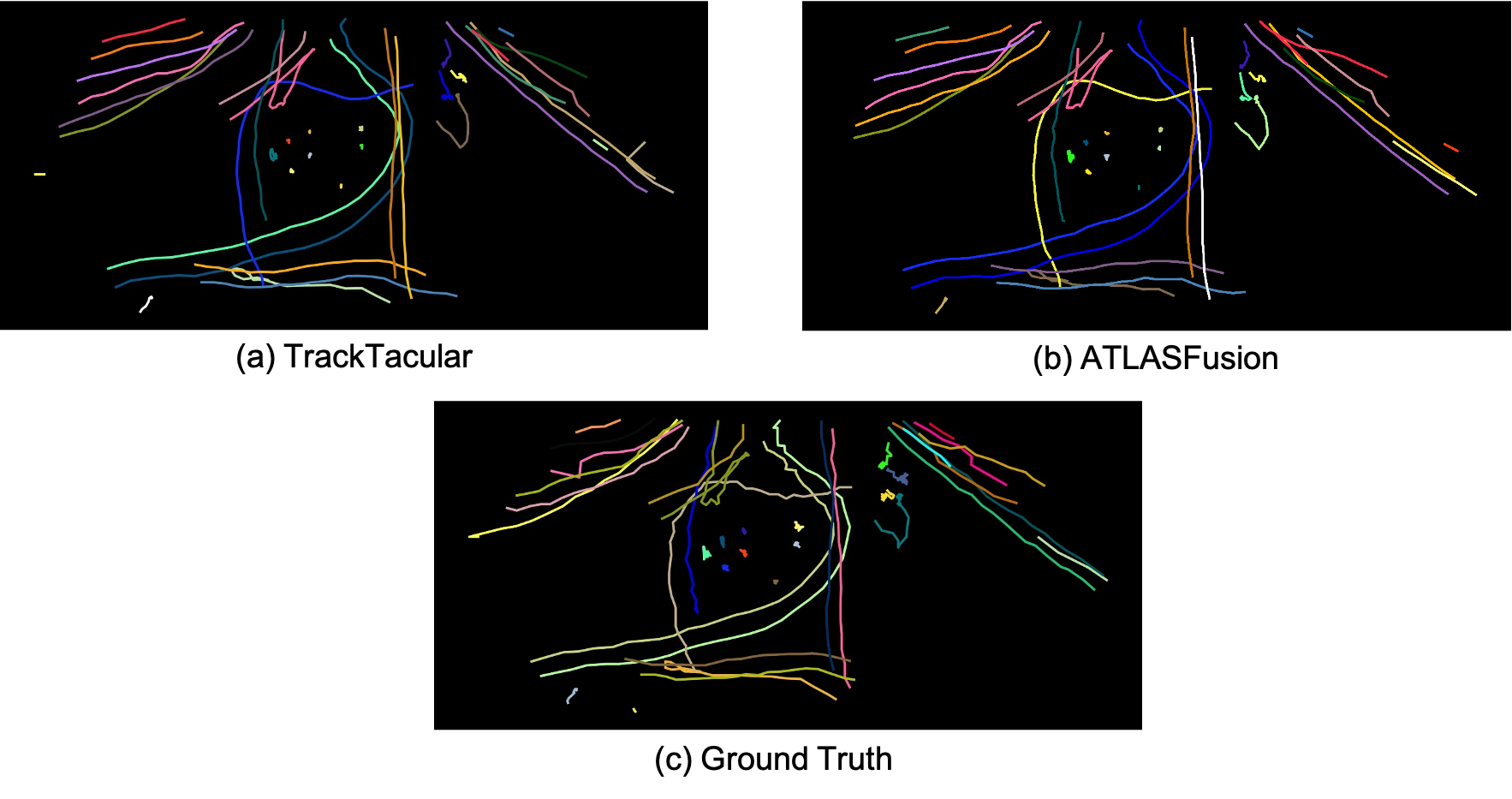}
  \caption{Qualitative comparison of tracking trajectories on the WildTrack dataset. (a) TrackTacular shows several ID switches and fragmented tracks (e.g., yellow and blue lines). (b) Our ATLASFusion produces more consistent and longer trajectories, successfully tracking objects across the scene. (c) Ground Truth for reference.
  }
  \label{fig:plot_mota}
\end{figure}

To further evaluate the contribution of each component, an ablation study was conducted on WildTrack and MultiViewX, as shown in \cref{tab:model-ablation-wild}. Three components were evaluated both individually and in combination: SPT, Weighted Aggregation (WA), and Per-View BEV Supervision (PV Loss).
In \cref{tab:model-ablation-wild}, adding SPT improved all metrics, notably boosting MODP from 76.2\% to 79.5\%. This result indicates that avoiding unnatural interpolation by projecting only valid features leads to more precise object localization, which is directly reflected in the MODP score.
Adding Density-Aware Weighting (+WA) further improved performance, especially on tracking metrics such as IDF1 and MOTA. By assigning higher confidence to features from nearby objects, distortions from distant, low-resolution regions were effectively mitigated. The resulting more reliable fused representation induced more stable tracking with fewer identity mismatches.
Per-View BEV Supervision (+PV Loss) provided the largest boost to IDF1 (95.9\%). This result indicates that directly supervising each camera's BEV feature before fusion improves the information available to the tracker. The higher IDF1 is consistent with improved identity preservation, although this loss does not explicitly couple different views.
On MultiViewX in \cref{tab:model-ablation-wild}, SPT improved localization precision, but the overall MODA remained nearly unchanged because the MultiViewX dataset is synthetic and already close to performance saturation.
Weight Aggregation reduced minor false positives and produced smoother feature fusion across cameras.
PV Loss directly supervised each view, leading to a slight gain in IDF1.



\cref{fig:plot_mota} provides a qualitative comparison of tracking trajectories on the WildTrack dataset, visually confirming the effectiveness of the proposed method.
The result produced by TrackTacular, shown in \cref{fig:plot_mota}(a), exhibits several noticeable tracking errors.
TrackTacular incorrectly assigns the same identity to nearby pedestrians on the left side, while ID switches are observed on the right. Several trajectories also break or bend unnaturally around the crossing area.
In contrast, trajectories generated by ATLASFusion, shown in \cref{fig:plot_mota}(b), are more stable and consistent, maintaining correct identities and continuous trajectories even in crowded intersections. The result closely mirrors the continuity of the ground truth shown in \cref{fig:plot_mota}(c).

The tracking results produced by ATLASFusion are longer with fewer fragments and color changes, providing clear visual evidence of improved identity preservation.
This visual improvement is attributed to the robust feature fusion mechanism. By mitigating feature distortions through sparse projection and density-aware weighting, ATLASFusion maintains a more reliable representation of each object, achieving more robust multi-object tracking.

\subsection{Robustness and Sensitivity Analysis}

To evaluate the robustness of the proposed sparse BEV projection, ATLASFusion was tested under two practical perturbation scenarios: calibration noise and reduced input resolution.

\subsubsection{\textbf{Calibration Noise.}}

Robustness to imprecise camera parameters was assessed by injecting Gaussian noise into the camera extrinsics (rotation $r^\circ$, translation $t$ m), following the calibration uncertainty reported for the WildTrack dataset.
As reported in \cref{tab:noise}, under moderate noise ($r=0.5^\circ$, $t=0.05$ m), TrackTacular exhibited a 41.6\% absolute drop in MODA, whereas ATLASFusion degraded by only 14.9\%.
Under severe noise ($r=2.0^\circ$, $t=0.2$ m), TrackTacular collapsed entirely (MODA 0.0\%), while ATLASFusion retained 40.1\% MODA.
These results indicate that SPT, which bypasses dense interpolation in favor of direct geometric projection, is inherently more resilient to calibration inaccuracies.

\begin{table}[t]
  \centering
  \setlength{\tabcolsep}{3.0pt}
  \caption{Robustness under calibration noise on WildTrack.}
  \scalebox{0.9}{
    \begin{tabular}{l|cc|cc}
      \toprule
      Noise Level                & \multicolumn{2}{c|}{TrackTacular} & \multicolumn{2}{c}{\textbf{Ours}}                                 \\
                                 & MODA                              & MODP                              & MODA          & MODP          \\
      \midrule
      Low ($0.5^\circ$, $0.05$m) & 50.5                              & 64.8                              & \textbf{78.7} & \textbf{73.9} \\
      High ($2.0^\circ$, $0.2$m) & 0.0                               & 46.4                              & \textbf{40.1} & \textbf{58.8} \\
      \bottomrule
    \end{tabular}}
  \label{tab:noise}
\end{table}

\subsubsection{\textbf{Resolution Sensitivity.}}

Performance was further evaluated under halved input resolution ($720\!\times\!1280 \to 360\!\times\!640$).
As reported in \cref{tab:resolution},
MVTr collapsed (IDF1 2.9\%, MOTA 0.0\%) because its Deformable DETR backbone was unable to reliably detect the smaller objects produced at reduced resolution.
TrackTacular incurred a 2.7\% decrease in MODA (92.1 $\to$ 89.4).
ATLASFusion sustained high accuracy with only a 1.1\% decrease in MODA (93.6 $\to$ 92.5), confirming superior robustness to reduced information density.

\subsection{Computational Efficiency}

\cref{tab:efficiency} summarizes the computational cost of ATLASFusion and competing methods.
MVTr and CaMuViD exhibited substantially higher computational costs due to their Deformable Transformer and InternImage backbones, respectively.
ATLASFusion and TrackTacular share a lightweight CNN + CenterNet backbone.
ATLASFusion operated at 9.90 FPS with 1.46 GB of GPU memory, achieving comparable throughput to TrackTacular while surpassing it in both detection and tracking accuracy (cf.\ \cref{fig:intro}).
This minimal overhead is attributed to the shared lightweight backbone employed for both detection and tracking; only the sparse projection and density-aware weighting modules were added.

\begin{table}[t]
  \centering
  \begin{minipage}[t]{0.52\textwidth}
    \centering
    \caption{Performance under reduced resolution on WildTrack.}
    \label{tab:resolution}
    \scalebox{0.85}{
      \begin{tabular}{l|ccc}
        \toprule
        Method                      & IDF1                                                & MOTA                                                & MODA                                                \\
        \midrule
        MVTr~\cite{mvtr}            & 2.9\textcolor{purple}{\scriptsize($-$90.2)}         & 0.0\textcolor{purple}{\scriptsize($-$92.3)}         & -                                                   \\
        TrackTacular~\cite{tacular} & 94.4 \textcolor{red}{\scriptsize($-$0.9)}           & 90.0 \textcolor{red}{\scriptsize($-$1.8)}           & 89.4 \textcolor{red}{\scriptsize($-$2.7)}           \\
        \textbf{ATLASFusion}        & \textbf{95.8} \textcolor{blue}{\scriptsize($-$0.1)} & \textbf{92.3} \textcolor{blue}{\scriptsize($-$0.1)} & \textbf{92.5} \textcolor{blue}{\scriptsize($-$1.1)} \\
        \bottomrule
      \end{tabular}}
  \end{minipage}%
  \hfill
  \begin{minipage}[t]{0.46\textwidth}
    \centering
    \caption{Computational efficiency comparison.}
    \label{tab:efficiency}
    \scalebox{0.85}{
      \begin{tabular}{l|cc}
        \toprule
        Method                      & FPS$\uparrow$ & Mem$\downarrow$ \\
        \midrule
        CaMuViD~\cite{camuvid}      & 5.08          & 8.14GB          \\
        MVTr~\cite{mvtr}            & 2.15          & 1.39GB          \\
        TrackTacular~\cite{tacular} & 9.88          & 1.46GB          \\
        \textbf{ATLASFusion}        & \textbf{9.90} & \textbf{1.46GB} \\
        \bottomrule
      \end{tabular}}
  \end{minipage}
\end{table}

\section{Conclusion and Future Work}

This paper presented ATLASFusion, a framework that addresses the feature distortion and non-uniform density arising when multi-view features are projected into a single BEV space for robust spatio-temporal multi-view pedestrian tracking. The proposed approach integrates three key components: a sparse projection method to prevent interpolation artifacts, density-aware weighting to prioritize reliable features, and per-view BEV supervision to ensure that each view learns a robust representation before fusion.
Experimental results on two benchmarks validated the effectiveness of ATLASFusion. ATLASFusion improved MODP from 75.0\% to 89.2\% over the TrackTacular baseline on MultiViewX and achieved an IDF1 of 95.9\%, the highest among the compared methods on WildTrack.
These results demonstrate that the proposed approach successfully integrates multi-view information to achieve robust and accurate scene understanding, mitigating the tracking inconsistencies caused by projection distortions.

Despite these improvements, promising directions for future work remain.ATLASFusion retains the flat-ground assumption, and its confidence map models geometric observation density rather than occlusion or image quality. Enhancing computational efficiency for real-time applications and developing methods that operate without pre-calibrated camera parameters are key steps toward practical deployment. Future research will focus on these areas to bring robust multi-view tracking closer to real-world deployment.

\clearpage  

%
%
\bibliographystyle{splncs04}
\bibliography{main}
\end{document}